# The Informational Cost of Agency: A Bounded Measurement Layer for Deployed Reinforcement Learning

Wael Hafez
Semarx Research LLC
Alexandria, VA, USA.
w.hafez@semarx.com

Cameron Reid
Semarx Research LLC
Alexandria, VA, USA
cameron.reid@semarx.com

Amir Nazeri
Semarx Research LLC
Alexandria, VA, USA
amir.nazeri@semarx.com

**Abstract—**Deployed reinforcement learning systems lack a principled runtime reliability theory. We close this gap by introducing Bipredictability ($P$), a closed-form information-theoretic metric that quantifies how efficiently a closed-loop interaction converts uncertainty into shared predictability between agent and environment. $P$ admits a provable classical bound $P \leq 0.5$, derived from Shannon entropy subadditivity, and responsive agency necessarily suppresses $P$ below this ceiling — a structural prediction we term the informational cost of agency. Across 21 trained continuous-control agents, we confirm this prediction empirically at $P = 0.33 \pm 0.02$. The same suppression signature reproduces in language-model dialogue, convolutional vision systems, and classical mechanical baselines, indicating that $P$ captures a substrate-independent property of agentic interaction rather than an algorithm-specific artifact. The Information Digital Twin (IDT), a model-agnostic architecture that computes $P$ from the external interaction stream, detects 89.3% of coupling degradations against 44.0% for reward-based monitoring, with $4.4\times$ lower latency. $P$ provides the missing measurement layer for runtime reliability and closed-loop self-regulation in deployed autonomous systems.

## 1 Introduction

The deployment of reinforcement learning (RL) agents (Sutton & Barto, 2018) into real-world environments has outpaced the development of principled frameworks for their runtime monitoring and assurance. RL agents exhibit complex behaviors derived from a continuous observation–action–outcome loop, and once a policy is frozen and deployed, the agent–environment coupling that supported training-time performance can be disrupted by environmental shifts, sensor degradation, or actuator drift — often silently, and often before any reward-visible failure (Dulac-Arnold et al., 2021; Josifovski et al., 2024; Levine et al., 2020). The field's most impressive demonstrations (Kaufmann et al., 2023; Radosavovic et al., 2023; Tang et al., 2024) were achieved without any principled runtime measure of whether the deployed policy remains coupled to its environment. Current reliability is managed through paradigms that address pieces of the problem but not the gap itself: out-of-distribution detection (Liang et al., 2018; Koh et al., 2021) and concept-drift methods (Greco et al., 2025) flag inputs that look unfamiliar; reward-based monitoring flags outcomes that look bad. Both are reactive, both require degradation to accumulate before deviating from baseline, and neither measures the closed-loop interaction itself. Model-internal diagnostics, where they exist, are frequently unavailable in third-party or black-box deployments.

We introduce Bipredictability ($P$) as a substrate-independent metric that quantifies the fraction of the total observation–action–outcome uncertainty budget converted into shared predictability across the closed-loop interaction. Whereas a policy describes how the agent acts; $P$ describes how well the agent and environment remain coupled. This makes $P$ sensitive to coupling degradation regardless of whether the policy itself has changed. Unlike empowerment (Klyubin et al., 2005; Salge et al., 2014; Mohamed & Rezende, 2015) and active inference (Friston, 2009; Friston et al., 2017), which characterize the agent's potential influence or its drive to minimize prediction error, $P$ measures the realized coupling against the interaction's full uncertainty budget at runtime, and is structurally bounded by the geometry of that budget. We establish that all classical interactions are subject to a structural ceiling $P \leq 0.5$, derived from Shannon entropy subadditivity (Shannon, 1948). The transition from passive coupling to responsive agency — where actions depend on observations — necessarily suppresses $P$ below this ceiling, a phenomenon we term the informational cost of agency.

Our primary contribution is the demonstration that this cost is not an artifact of any specific implementation but a structural signature of agency. Empirical validation across 21 continuous-control agents trained with SAC and PPO in the MuJoCo HalfCheetah environment yields a consistent baseline of $P = 0.33 \pm 0.02$. This finding is corroborated by independent measurements across substrates that share no dynamical structure with continuous-control RL: multi-turn language-model dialogue (Hafez & Nazeri, 2026), convolutional vision systems (Nazeri & Hafez, 2025), and a non-agentic classical mechanical baseline (Hafez et al., 2026). The framework extends theoretically to quantum-entangled systems, where the classical bound generalises to $P \leq 1$ under unitary evolution (Hafez et al., 2026). The convergence of agentic systems toward $P \approx 1/3$ across these heterogeneous settings indicates that $P$ functions as a structural property of the closed-loop interaction itself, not a property of any specific algorithm.

To operationalize this measurement for autonomous systems engineering, we present the Information Digital Twin (IDT), a model-agnostic architecture that computes $P$ from the external interaction stream without access to policy weights, internal activations, or reward signals. Across 168 perturbation trials spanning eight perturbation types and two policy architectures, IDT-based monitoring detected 89.3% of coupling degradations versus 44.0% for reward-based monitoring, with 4.4 × lower median latency.

The empirical results presented here rest on two foundations: a theoretical framework derived from first principles in a companion paper (Hafez et al., 2026), and an architecture — the IDT — that operationalizes the framework as a real-time measurement instrument. Section 2 summarizes the framework. Section 3 describes the IDT architecture. Section 4 presents results, including the cross-substrate convergence. Section 5 discusses implications for runtime reliability and self-regulation. Section 6 details experimental methods.

## 2 Bipredictability framework

### 2.1 Uncertainty resolution requires an information-theoretic framework

Deployment monitoring is fundamentally a question about uncertainty resolution: whether observations and actions reduce uncertainty about outcomes, and whether outcomes constrain what the agent must have done. Information theory operationalizes this directly — entropy quantifies uncertainty, mutual information quantifies its resolution.

The minimal sufficient description of a closed-loop interaction requires three elements: what the agent perceives ($S$), what it does ($A$), and what results ($S'$). Information theory maps naturally onto this structure. Each variable carries uncertainty quantified by its Shannon entropy — $H(S), H(A)$, and $H(S')$ — and the degree to which they resolve each other's uncertainty is captured by mutual information. Figure 1 makes this geometry explicit: each circle represents the entropy of one variable, their central overlap represents the shared predictability $MI(S, A;\ S')$ realized across the full loop, and the total area represents the complete uncertainty budget $C$ available in the interaction. The deployment monitoring question then has a precise geometric form: what fraction of the total area is the overlap?

Conditional entropy quantifies what remains unresolved: $H(S'|S, A)$ is the region of $S'$ falling outside its overlap with $(S, A)$ in Figure 1, and $H(S, A|S')$ is the region of $(S, A)$ falling outside its overlap with $S'$. When the overlap is large relative to the total area, each side of the interaction reliably predicts the other; when it is small, the loop is informationally loose.

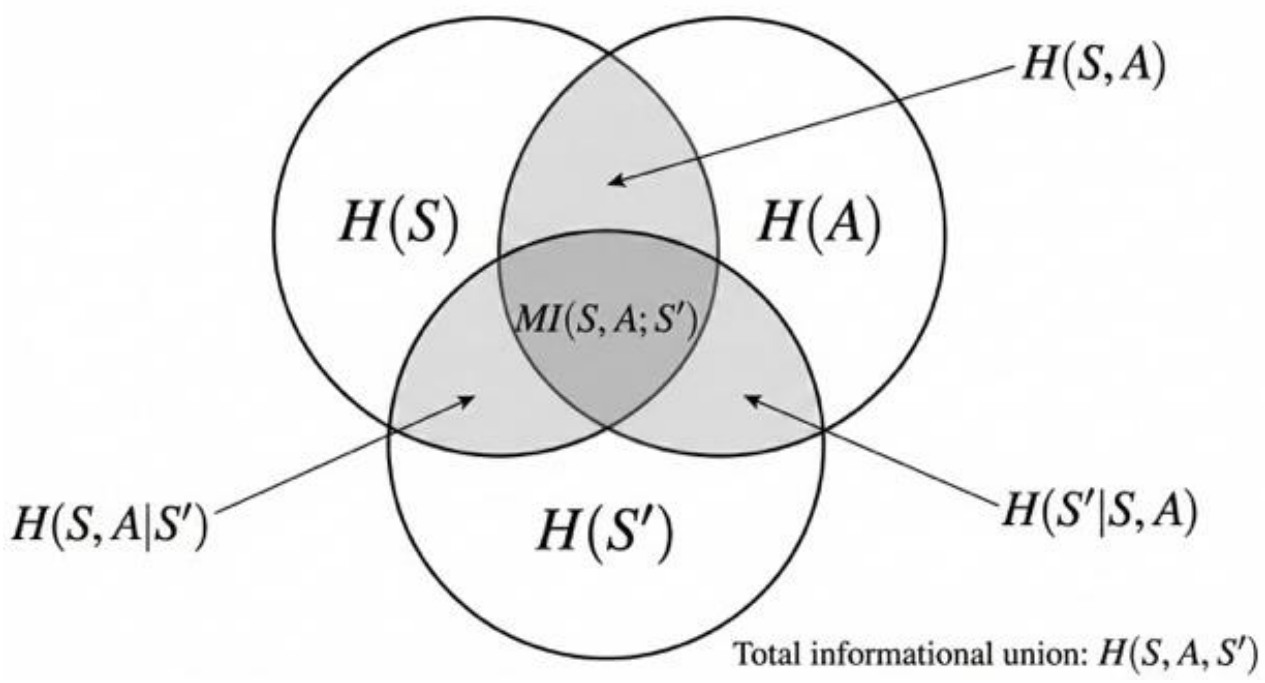

**Figure 1.** Information-theoretic structure of the observation–action–outcome interaction. Each circle represents the entropy of one variable: observations $H(S)$, actions $H(A)$, and outcomes $H(S')$. The central overlap is the mutual information $MI(S, A; S')$; non-overlapping regions correspond to conditional entropies. *Note on visual representation:* While the central overlap is labeled $MI(S, A; S')$ for clarity, it serves as a visual proxy for the joint mutual information; mathematically, $MI(S, A; S')$ corresponds to the entire region of $S'$ shared with the $(S, A)$ union. Bipredictability ($P$) is defined as the ratio of this shared information to the total informational union $H(S, A, S')$.

This bidirectional uncertainty resolution is also what makes agency informationally costly: when actions depend on observations — as they must in any responsive agent — part of the uncertainty budget is consumed by the observation–action dependency before any outcome is realized, reducing the fraction available for shared predictability. We term this the informational cost of agency.

### 2.2 Bipredictability measures interaction efficiency against a provable classical bound

The classical bound $P \leq 0.5$ and the structural suppression below this ceiling under observation-dependent action — the *informational cost of agency* — are derived from first principles in (Hafez et al., 2026). We restate the bound here for self-containment and test both predictions empirically in Section 4.

The geometry of Figure 1 suggests a natural measure of interaction efficiency: the ratio of the central overlap to the total area. Formally, we define Bipredictability $P$ as the fraction of the total uncertainty budget $C = H(S) + H(A) + H(S')$ converted into shared predictability across the full loop:

$$P = \frac{MI(S, A; S')}{H(S) + H(A) + H(S')} \qquad (1)$$

At $P = 0$ the observation–action pair and outcome are statistically independent. Higher values indicate tighter coupling. Because $P$ is normalized by the full uncertainty budget rather than any single variable's entropy, it remains meaningful when interaction scope changes — a property pairwise measures such as normalized mutual information do not share. $P$ is related to but distinct from empowerment (Klyubin et al., 2005; Salge et al., 2014; Mohamed & Rezende, 2015), which quantifies the channel capacity $I(A; S'|S)$ of an agent's action channel and is typically maximized during training as an intrinsic motivation signal. $P$ differs in three respects: it is normalized by the full interaction uncertainty budget rather than measured on a single channel, it is structurally bounded above by ½ rather than unbounded, and it is defined for runtime measurement of realized coupling rather than policy optimization.

A strict upper bound follows directly from Shannon entropy (Shannon, 1948; Cover & Thomas, 2006). Mutual information cannot exceed the entropy of either side, and by subadditivity:

$$MI(S,A;S') \leq \min\big(H(S,A),H(S')\big) \leq \min\big(H(S)+H(A),H(S')\big) \qquad (2)$$

Let the total uncertainty budget be $C = H(S) + H(A) + H(S')$. Under the constraint that $C$ is fixed, the ratio $MI/C$ is maximised when both $H(S) + H(A)$ and $H(S')$ are as large as possible while mutually constraining each other. The maximum occurs when $H(S) + H(A) = H(S') = C/2$, yielding:

$$P = \frac{MI}{C} \leq \frac{\frac{C}{2}}{\mathrm{C}} = 0.5 \qquad (3)$$

Thus, no classical interaction loop can convert more than half of its total uncertainty into shared predictability. Responsive agents, in which actions genuinely depend on observations, necessarily consume some of the budget in the $S \rightarrow A$ dependency, suppressing $P$ below this ceiling. The $1/3$ baseline observed across trained agents empirically quantifies this structural cost.

This bound is structural — it holds for any classical system representable over discrete variables, independent of domain, task, or agent architecture. The full derivation and saturation conditions are given in Hafez et al. (2026). The mechanism is that responsive action introduces a statistical dependency between $S$ and $A$ that consumes part of the uncertainty budget without contributing to shared predictability across the loop. The empirical baseline reported in Section 4.1 — $P = 0.33 \pm 0.02$ across trained agents — confirms this prediction.

### 2.3 Bipredictability decomposes into forward, backward, and asymmetry components

$P$ measures overall interaction efficiency but detecting that coupling has degraded is only the first requirement for deployment monitoring — understanding *where* the degradation originates is what enables intervention. A drop in $P$ could reflect growing environmental unpredictability, loss of action legibility, or both simultaneously. To distinguish these, we analyze the two directions of uncertainty flow within the loop.

- Forward predictive uncertainty measures how much the outcome remains unresolved after the observation and action are known:

  $$Hf = H(S' \mid S,A) \qquad (4)$$

  When *Hf* increases, the environment responds in ways the agent cannot anticipate — outcomes are no longer well-constrained by the state–action pair.

- Backward predictive uncertainty measures how much the observation–action pair remains unresolved after the outcome is observed:

  $$Hb = H(S,A \mid S') \qquad (5)$$

  When *Hb* increases, many distinct observation–action pairs lead to the same outcome — the agent's internal distinctions are invisible from the environment's side, indicating loss of action legibility.

Their difference defines predictive asymmetry:

$$\Delta H = Hf - Hb \qquad (6)$$

In passive systems without responsive action, uncertainties are balanced and $\Delta H \approx 0$ — consistent with double pendulum results reported in Hafez et al. (2026). Responsive agency disrupts this balance, inducing persistent asymmetry whose sign reveals the dominant source of coupling failure: environment-side disruptions drive $Hf$, while agent-side degradation drives $Hb$. Together, $P, Hf, Hb$, and $\Delta H$ form a diagnostic set — $P$ signals that something has changed, while the components localize where.

## 3 The Information Digital Twin (IDT): Real-time monitoring of Bipredictability

Section 2 showed that $P$ captures the shared information structure of the interaction loop, while its components, $Hf$ , $Hb$ and $\Delta H$, localize predictive failures. To serve as a deployment monitoring signal, these quantities must be computed online from the interaction stream, without access to internal model parameters or reward signals. We introduce the Information Digital Twin (IDT) (Fig. 2), an auxiliary architecture that runs alongside the deployed agent and computes $P$ and its components directly from the $(S, A, S')$ stream. The IDT adopts the digital twin paradigm (Grieves & Vickers, 2017; Tao, Zhang & Nee, 2019) — a real-time computational counterpart of a physical or computational system — and specializes it for monitoring information-theoretic structure rather than physical state.

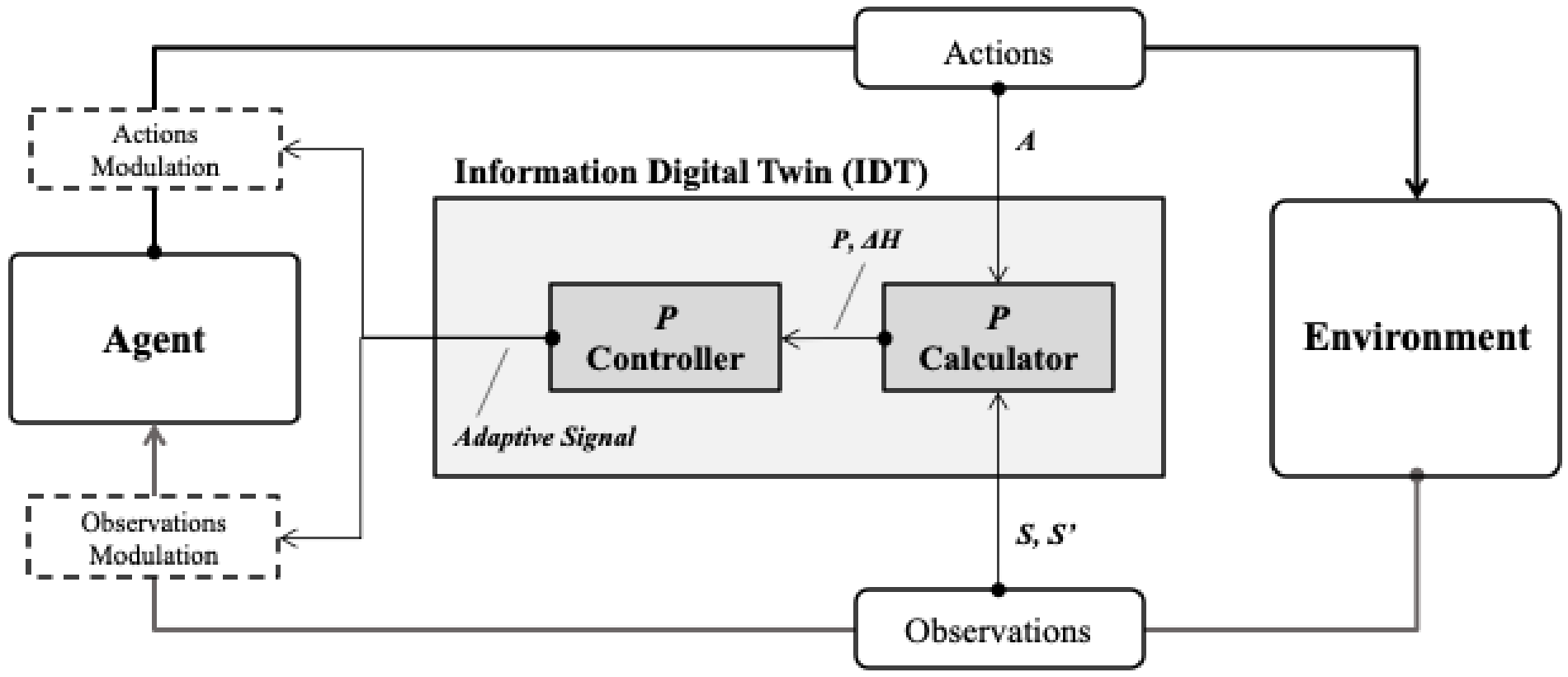

**Figure 2.** Information Digital Twin (IDT) architecture. The IDT operates alongside the agent–environment loop, receiving copies of observations $(S, S')$ and actions $(A)$. The $P$ Calculator calculates Bipredictability $P$ and predictive asymmetry $\Delta H$ from the interaction stream. The $P$ Controller detects statistical deviations from baseline coupling. Dashed boxes indicate architecturally specified modulation pathways — observation modulation and action modulation — that are not experimentally validated in this work. The Adaptive Signal path connects controller output to the modulation interfaces.

The IDT operates in three stages: online metric estimation (3.1), statistical detection of deviations from the learned baseline (3.2), and a modulation stage—architecturally specified but not yet experimentally validated—that would adapt the agent's observation and action interfaces in response to sustained deviations (3.3).

### 3.1 $P$ is computed online without access to model internals

The IDT treats the deployed agent as a black box: it observes only the externally visible interaction stream and requires no access to policy weights, internal activations, or reward signals. At each time step, the agent selects an action $A$ given its current observation $S$, the environment transitions, and a new observation $S'$ is produced. The IDT intercepts copies of this $(S, A, S')$ tuple and estimates $P$ and its diagnostic components through the following pipeline:

At each timestep, the IDT records the $(S, A, S')$ tuple, accumulates tuples over a sliding window, computes the marginal entropies, joint mutual information, and conditional entropies from the windowed distributions, and derives $P$ and $\Delta H$ for that window. The pipeline repeats at each window stride producing a time series of $P$ and $\Delta H$ values that track coupling integrity as the agent operates. The pipeline itself is general: it applies to any system expressible as an $(S, A, S')$ interaction loop, with the estimation fidelity governed by the windowing and discretization parameters.

### 3.2 Coupling degradation is detected via multi-channel statistical deviation

The time series produced by the estimation pipeline provides a continuous readout of coupling integrity, but raw values of $P$ and $\Delta H$ are not directly interpretable without a reference. The $P$ controller (Fig. 2) establishes this reference during an initial calibration period under nominal operation by estimating the baseline mean $\mu$ and standard deviation $\sigma$ for each metric. Subsequent values are evaluated against this learned baseline using the following protocol:

The $P$ controller calibrates baseline μ and σ for each metric over a pre-perturbation window, then flags any post-perturbation window in which any metric $- P, \Delta H, Hf$, or $Hb$ — deviates beyond $\pm 3\sigma$. The union of channels constitutes the IDT detection signal; trials with no threshold crossing are recorded as undetected.

The pattern of which metrics deviate, and in which direction, carries diagnostic information that differs across perturbation types. These diagnostic signatures are characterized in Section 4.4; operationalizing them as an automated attribution mechanism is future work.

The $\pm 3\sigma$ threshold is used because deviations in either direction indicate issues: in general terms, a decrease in $P$ shows decoherence, while an increase indicates rigidity. Using four metrics improves detection, since different perturbations affect components differently.

### 3.3 $P$ provides the prerequisite signal for closed-loop self-regulation

A complete reliability architecture would not stop at detection and attribution. It would close the loop, using the measured deviation as feedback to adjust the agent's coupling with the environment in real time. The IDT specifies this modulation layer as a controller that receives $(P, Hf, Hb, \Delta H)$ deviations from the learned baseline and emits bandwidth adjustments at the interaction interface — observation filtering when deviations are $Hf$-dominated, action damping when deviations are $Hb$-dominated — without altering policy weights, preserving learned behavior while mitigating coupling degradation. The interface is defined; the control law is not. We do not claim experimental validation of this layer in the present paper. We claim that the measurement and attribution layers together establish the prerequisite signal that any such control law would require — the analog, in deployment-reliability terms, of the measurable channel capacity that made principled communication-system design possible. Demonstrating closed-loop self-regulation built on this foundation is the natural next step.

## 4 Results

### 4.1 Trained agents exhibit stable coupling below the classical bound

Before evaluating detection performance, we characterize the informational baseline of nominal agent–environment interaction. Under unperturbed operation, the 21 trained agents exhibit a mean Bipredictability of $P = 0.33 \pm 0.02$ and a mean predictive asymmetry of $\Delta H = -0.56 \pm 0.22$, stable across the pre-perturbation episodes of all 168 evaluation runs. The baseline $P \approx 1/3$ lies well below the classical upper bound of 0.5 (Hafez et al., 2026), confirming the theoretically predicted informational cost of agency: responsive, observation-dependent action necessarily suppresses interaction efficiency below the classical ceiling. This value is consistent with the agency-cost regime observed across other substrates (Section 4.5), suggesting the suppression reflects a structural property of observation-dependent action rather than a HalfCheetah-specific artifact. The negative $\Delta H$ shows that forward prediction (anticipating $S'$ from $S$ and $A$) carries greater residual uncertainty than backward inference (recovering $S$ and $A$ from $S'$), reflecting the asymmetry of active control: committed actions constrain backward inference more strongly than stochastic environmental responses constrain forward prediction. These baselines define the reference for perturbation-induced deviations. Each seed's baseline mean and standard deviation are used to calibrate the $\pm 3\sigma$ detection protocol.

### 4.2 IDT-based monitoring detects twice the perturbations that reward signals miss

We evaluate whether monitoring Bipredictability from the interaction stream detects coupling degradation missed by reward-based monitoring. The $\pm 3\sigma$ detection protocol is applied identically to IDT metrics ($P, \Delta H, Hf, Hb$) and windowed episode reward across all 168 perturbation trials.

IDT-based monitoring detected $89.3 \pm 15.1\%$ of perturbations, compared with $44.0 \pm 26.1\%$ for reward-based detection (paired t-test, $n = 21$ seeds; $t = 7.95$, $p < 10^{-6}$, $d = 1.73$). Detection rates were computed per seed as the proportion of perturbation trials detected and averaged across seeds. The improvement arises from two factors: IDT metrics respond to changes in interaction structure that may not immediately affect cumulative reward, and the union of four diagnostic channels provides broader coverage than a single reward signal. A detection is registered when any metric exceeds its $\pm 3\sigma$ threshold; different perturbations activate different subsets of channels, but collectively they capture most conditions tested. Reward-based monitoring remains effective for perturbations that cause large, sustained performance drops. However, perturbations that degrade coupling without immediate reward collapse (e.g., moderate observation noise partially compensated by the policy) often go undetected by reward while producing clear deviations in $P$ or its components. This "silent degradation" regime motivates the proposed framework.

### 4.3 The IDT detects coupling degradation 4.4× faster than reward

Beyond detecting more perturbations, timely detection is critical for deployment monitoring, as early warning enables intervention before irreversible performance degradation. Detection latency is measured as the number of observation windows between perturbation onset (episode 15) and the first window in which any metric exceeds its $\pm 3\sigma$ threshold.

Across all 168 perturbation trials, IDT-based monitoring achieved a median detection latency of 42 windows, compared with 184 windows for reward-based detection—a $4.4\times$ improvement. This gap reflects a fundamental signal difference: $P$ responds to structural changes in the interaction at the transition level, whereas reward aggregates effects over episodes and requires degradation to accumulate before deviating. In deployment terms, earlier detection provides a larger window for corrective action, including operator alerts, fallback policies, or—within a future closed-loop architecture—reflexive modulation.

### 4.4 The diagnostic decomposition provides complementary detection channels

The IDT's detection advantage over reward arises from the complementary sensitivity of multiple information-theoretic channels rather than any single metric. Table 1 reports the individual detection rate and median latency for each monitored quantity, their union (IDT), and reward. No single metric dominates: individual detection rates cluster between $69 - 73\%$, exceeding reward-based detection (44.0%) but falling well short of the union (89.3%).

**TABLE 1.** Per-metric detection rate and median latency across 168 perturbation trials. IDT (union) registers detection when any metric exceeds $\pm 3\sigma$. $P$ = Bipredictability; $Hf = H(S'|S,A)$; $Hb = H(S,A|S')$; $\Delta H = Hf - Hb$. Reward uses the same ±3σ protocol on windowed episodic return.

| Metric | Detection Rate (%) | Median Latency (windows) |
|---|---|---|
| IDT (union) | 89.3 | 42 |
| $P$ | 73.2 | 74 |
| $Hf$ | 70.8 | 69 |
| $Hb$ | 69.6 | 75 |
| $\Delta H$ | 69.0 | 67 |
| Reward | 44.0 | 184 |

A similar pattern holds for latency, with individual medians between 67–75 windows versus 42 for the union. This gap—approximately 16 percentage points in coverage and 25 windows in latency—demonstrates that each channel detects perturbations missed by the others; the metrics are complementary, not redundant.

This complementarity is further reflected in per-algorithm effect sizes. For SAC agents, information-theoretic metrics are approximately twice as sensitive as reward (mean Cohen's $d = 0.93 - 1.02$ vs. $0.60$), whereas PPO agents exhibit comparable sensitivity across all metrics ($d \approx 0.5$). Nevertheless, the IDT outperforms reward-based detection for both algorithms—through stronger individual signals for SAC and through the union of channels for PPO, where individual metrics alone are insufficient.

### 4.5 The agency-cost signature reproduces across substrates

The classical bound $P \leq 0.5$ is a structural property of Shannon entropy and applies to any closed-loop interaction with finite state spaces. The agency-cost suppression toward $P \approx 0.3$ is empirical, and a single-domain confirmation cannot distinguish a structural regularity from a HalfCheetah-specific artifact. We therefore situate the present results within the broader cross-substrate evidence.

Three independent validations, conducted under the same definition of $P$ and the same protocol of computing it from external interaction streams, report converging values. In multi-turn language-model dialogue across three frontier models, $P = 0.275 \pm 0.029$, with $100\%$ perturbation detection across contradiction, topic-shift, and non-sequitur injections (Hafez & Nazeri, 2026). In convolutional vision systems under input perturbation, the entropy and mutual-information components from which $P$ is constructed exhibit the same structured sensitivity, yielding $90\%$ noise-detection accuracy with $0\%$ false positives (Nazeri & Hafez, 2025). As a non-agentic control, a double-pendulum classical mechanical system without observation-dependent action sits at $P \approx 0.48$, near the theoretical ceiling — the regime predicted for systems where the agency cost does not apply (Hafez et al., 2026).

These are not commensurate experiments. They differ in dimensionality, dynamics, and computation. They share only the definition of $P$ and the presence or absence of observation-dependent action. That such heterogeneous systems converge on the same regime structure — agency-bearing systems suppressed to roughly one-third of the uncertainty budget, agency-free systems near the classical ceiling — is the strongest available evidence that the signature observed here is substrate-independent rather than HalfCheetah-specific. A formal cross-substrate meta-analysis is beyond the scope of this paper, which focuses on the deepest single-domain validation.

## 5 Discussion

The central finding of this work is that the information structure of the observation–action–outcome loop carries a deployment-relevant signal that reward-based monitoring cannot access. Across 168 perturbation trials in continuous control, IDT-based monitoring detected $89.3\%$ of coupling degradations versus $44.0\%$ for reward, with $4.4 \times$ lower median latency. This advantage is structural rather than incidental. Reward aggregates task outcomes over episodes and requires degradation to accumulate before deviating; $P$ measures the fraction of the available uncertainty budget converted into shared predictability at the transition level, responding to coupling changes that may never manifest as reward collapse.

The gap between the empirical baseline $P \approx 1/3$ and the classical ceiling $P \leq 0.5$ is not a performance limitation — it is the informational cost of agency, the structural consequence of observation-dependent action that makes closed-loop control possible. This interpretation is supported by the convergence of the same regime structure across radically different substrates, each measured with its own discretisation and estimation pipeline. A classical double pendulum, convolutional neural networks, and multi-turn language models all exhibit the same qualitative ordering: agentic systems suppressed to roughly one-third of the uncertainty budget, non-agentic systems near the classical ceiling. The framework extends theoretically beyond the classical regime — quantum-entangled systems are predicted to reach $P \leq 1$ under unitary evolution — but this extension is mathematical rather than empirical. That such heterogeneous systems share only the definition of $P$ and the presence or absence of observation-dependent

action — yet yield a consistent signature — strongly suggests that the *nature* of what $P$ captures is invariant to the details of the estimation pipeline. Binning granularity and window length shift absolute scalar values, but the regime classification relative to the bound, and the separation between agentic and non-agentic dynamics, appear robust.

It is important to distinguish the metric from its operationalization. The theoretical bound and the qualitative suppression signature are proven or observed across multiple domains; the specific detection rates (89.3%) and latency improvements (4.4 ×) reported here are properties of a particular IDT configuration (three equal-width bins, 300-step windows, $\pm 3\sigma$ thresholds) applied to a single continuous-control environment. A formal sensitivity analysis across bin counts, window sizes, and threshold choices remains future work. However, given that the underlying metric is structurally robust, the IDT's detection advantage over reward — which arises from measuring coupling efficiency directly at the transition level — is unlikely to be an artefact of fine-tuning those parameters. The cross-substrate evidence provides strong independent support that the interpretive power of $P$ does not hinge on a specific discretisation.

Several limitations should be acknowledged. The full IDT pipeline was validated in a single continuous-control environment (HalfCheetah); multi-environment RL validation, while strongly indicated by the cross-substrate evidence, is not yet performed within the RL domain itself. The detection results are reported without a dedicated false-positive analysis on held-out unperturbed data, although the calibration on 14 unperturbed episodes per seed provides an implicit baseline stability estimate. Most importantly, this paper validates measurement and attribution; the modulation layer of the IDT, which would close the loop from monitoring to self-regulation, is architecturally specified but not experimentally demonstrated. What this paper establishes is the prerequisite without which any such control law would be impossible: a real-time, model-agnostic, structurally bounded measurement of agent–environment coupling.

The cost of operating without such an instrument is already visible — training compute redone whenever deployment conditions shift, reliability strategies that detect failures only after they occur, and an absence of runtime guarantees that has become the binding constraint on real-world autonomous deployment. Bipredictability does not solve these problems. It provides the measurement layer required to begin solving them.

## 6 Methods

### 6.1 Online estimation of $P$ from interaction streams

Bipredictability is defined over discrete random variables. Real interaction streams are continuous, requiring discretization before $P$ can be estimated. We apply z-score normalization and map continuous values into discrete bins, converting the raw $(S, A, S')$ stream into a sequence of transition symbols from which empirical joint and marginal distributions are constructed. Distributions are estimated over sliding windows of W transitions advanced by stride δ, producing a time series of $P, Hf, Hb,$ and $\Delta H$.

A distinction worth making explicit: $P$ is defined analytically as the ratio $MI(S, A;\ S')\ /\ [H(S)\ +\ H(A)\ +\ H(S')]$ and its bound $P\ \leq\ 0.5$ is a mathematical consequence of Shannon entropy subadditivity, independent of any discretization resolution, binning strategy, or window size. The bound is a property of the joint distribution of the interaction loop; estimation choices affect only how that distribution is approximated. What requires empirical investigation is not the bound itself, but whether the agency-cost suppression — the regime where agentic systems settle near $P\ \approx\ 1/3$ — is robust to the estimation pipeline.

The cross-substrate evidence summarized in Section 4.5 bears on this directly. The same regime structure appears in continuous-control RL agents (this work), multi-turn LLM dialogue (Hafez & Nazeri, 2026), and the entropy and mutual-information components of convolutional vision systems (Nazeri & Hafez, 2025), while a non-agentic classical pendulum sits near the theoretical ceiling (Hafez et al., 2026). These systems differ in dimensionality, dynamics, and discretization scheme — high-dimensional continuous state in RL and CNNs, naturally discrete tokens in language, low-dimensional continuous mechanics in the pendulum. That the regime separation between agentic and non-agentic dynamics persists across these heterogeneous estimation pipelines indicates that the metric's interpretive power does not hinge on a particular operationalization.

For the HalfCheetah experiments specifically, we fixed the discretization at three equal-width bins and a window of 300 timesteps after confirming that these choices produce stable entropy estimates. A formal sensitivity analysis across bin counts and window lengths in the RL setting is left for future work.

### 6.2 Discretisation, bounds, and estimation robustness

The discretization pipeline interacts with the theoretical bound in a way worth making explicit. The bound $P \leq 0.5$ is a property of the joint probability distribution; discretization produces an empirical distribution from which we compute an estimate $P$. By the data-processing inequality, coarse-graining cannot increase mutual information, so the mutual information estimated from discretized variables is a lower bound on the continuous mutual information. The sum of marginal entropies, by contrast, can shift in either direction with binning. Consequently, $P$ is not guaranteed to be a strict lower bound on the continuous $P$ for every binning scheme, but the ceiling $P \leq 0.5$ remains valid for the discretized joint distribution itself, since the derivation in Section 2.2 applies to any discrete-variable representation. Any apparent violation of the $0.5$ ceiling in an empirical estimate would therefore indicate an estimation artifact — typically from insufficient window size — rather than a failure of the bound.

The observed baseline $P \approx 1/3$ across our experiments lies well below $0.5$, indicating that the suppression is genuine rather than a binning artifact. Moreover, when the same discretization scheme is applied to both agentic and non-agentic systems, any systematic bias from coarse-graining applies symmetrically to both and does not affect the regime classification. This is why the cross-substrate signature — $P \approx 1/3$ with responsive agency, $P \leq 0.5$ without — survives across radically different discretization choices.

### 6.3 Environment and agents

We evaluate our framework on MuJoCo HalfCheetah-v4 (Todorov et al., 2012), a continuous-control benchmark with 17-dimensional observations and 6-dimensional torque actions. The task is to maximize forward velocity under control costs. HalfCheetah provides sufficient complexity for nontrivial information-theoretic estimation while remaining interpretable. Agents are trained using Soft Actor-Critic (SAC) (Haarnoja et al., 2018) and Proximal Policy Optimization (PPO) (Schulman et al., 2017), implemented in Stable-Baselines-3 (Raffin et al., 2021) with default hyperparameters.

This pairing tests whether Bipredictability monitoring generalizes across policy architectures. SAC agents were trained to convergence ($\approx 10k - 17k\ returns$) over $\sim 3\ M$ steps, while PPO agents converged faster ($\approx 6k - 8k\ returns$) over $\sim 1.5\ M$ steps. Convergence was defined by stable returns over a sustained evaluation window, after which policies were frozen. Using algorithms with different performance levels tests whether $P$ captures interaction structure independent of raw task performance. In total, 11 SAC and 10 PPO seeds were evaluated, yielding 21 agents. Each evaluation comprised 50 episodes (50,000 steps each). Freezing policies ensures that all observed $P$ deviations reflect perturbation-induced coupling changes rather than ongoing learning. The perturbation suite is described next.

### 6.4 Perturbation design

To assess generalization across failure modes, we designed a perturbation suite spanning environment-side interaction changes and agent-side observation and action degradations, with type and severity varied to test detection sensitivity. Environment perturbations include gravity changes (one level) and external force impulses (two levels), while agent perturbations include observation noise (two levels) and action noise (three levels). The eight conditions are summarized in Table 2.

**TABLE 2.** Perturbation suite. Eight perturbation conditions spanning agent-side and environment-side changes applied to all 21 agents for 168 total trials.

| Category | Perturbation Type | Parameter |
|---|---|---|
| Agent | Actuator noise | 1% Gaussian noise on actions |
| Agent | Actuator noise | 3% Gaussian noise on actions |
| Agent | Actuator noise | 4% Gaussian noise on actions |
| Environment | External force | 5 N force applied to torso, x- axis |
| Environment | External force | 10 N force applied to torso, x-axis |
| Environment | Gravity | Gravity increased to 110% |
| Agent | Observation noise | 1% Gaussian noise on observations |
| Agent | Observation noise | 3% Gaussian noise on observations |

This suite does not exhaust the space of possible perturbations, which is effectively unbounded in continuous control settings. Instead, it samples qualitatively distinct failure mechanisms across multiple severity levels, supporting task-independent sensitivity rather than tuning to specific failure modes. Each perturbation is introduced at episode 15 of a 50-episode evaluation, yielding 14 unperturbed episodes for baseline calibration. All eight perturbation types are applied to each of the 21 agents, for a total of 168 trials. Information metric computation for these trials is described next.

### 6.5 Information metrics computation and comparison protocol

Computing information-theoretic quantities from the HalfCheetah interaction stream requires discretizing 40 continuous variables per timestep (17 state, 6 actions, 17 next-state). We apply a three-step procedure: z-score normalization, discretization into three equal-width bins, and grouping variables by body part (front leg, back leg, torso) following the environment's kinematic structure. Per-variable bins within each group are concatenated to form composite symbols for $(S, A, S')$. This grouping preserves embodiment structure while keeping the joint distribution tractable. The three-bin choice was empirically validated: four bins produced unreliable entropy estimates at available window sizes, while quantile-based binning yielded flat, uninformative metrics.

$P, Hf, Hb,$ and $\Delta H$ are computed over sliding windows of 300 timesteps with a stride of 50, yielding 991 windows per evaluation. All quantities use standard base-2 entropy formulas. For comparison, per-step reward is averaged over identical windows, and the same $\pm 3\sigma$ detection protocol is applied to both IDT metrics and windowed reward, ensuring matched temporal resolution and fair comparison.